\documentclass[aps,pra,twocolumn,10pt,groupedaddress,nofootinbib,superscriptaddress,notitlepage,showpacs,floatfix]{revtex4-2}
\usepackage{graphicx,graphics}
\usepackage{subfigure}
\usepackage{dcolumn}   
\usepackage{array}
\usepackage{amsthm}
\usepackage{amsmath}
\usepackage{amssymb}
\usepackage{amsfonts}
\usepackage{mathrsfs} 
\usepackage{bm}
\usepackage{braket} 
\usepackage{enumitem}
\usepackage{url}
\usepackage[pdfstartview=FitH,pdfencoding=auto]{hyperref}
\usepackage{bookmark}
\usepackage{pifont}
\hypersetup{
    colorlinks=true,        		
    linkcolor=blue,              	
    citecolor=blue,        			
    filecolor=blue,          		
    urlcolor=blue,              	
    runcolor=cyan
			}
\usepackage[pdftex]{color}
\usepackage{multirow}
\usepackage{CJK}
\usepackage{cancel}
\usepackage{ulem}
\usepackage[linesnumbered,ruled,vlined,noend]{algorithm2e}

\begin{document}
\begin{CJK*}{GB}{}
\title{Hybrid Quantum-Classical Neural Networks for Few-Shot Credit Risk
Assessment}
\author{Zheng-an Wang}
	\thanks{The authors contributed equally to this work.}
\affiliation{Beijing Key Laboratory of Fault-Tolerant Quantum Computing, Beijing Academy of Quantum Information Sciences, Beijing 100193, China}
\author{Yanbo J. Wang}
	\thanks{The authors contributed equally to this work.}
\affiliation{Longying Zhida (Beijing) Technology Co., Ltd., Beijing 100020, China}
\author{Jiachi Zhang}
	\thanks{The authors contributed equally to this work.}
\affiliation{Beijing National Laboratory for Condensed Matter Physics, Institute of Physics, Chinese Academy of Sciences, Beijing 100190, China}
\affiliation{School of Physical Sciences, University of Chinese Academy of Sciences, Beijing 100190, China}
\author{Qi Xu}
\affiliation{Longying Zhida (Beijing) Technology Co., Ltd., Beijing 100020, China}
\author{Yilun Zhao}
\affiliation{Beijing Key Laboratory of Fault-Tolerant Quantum Computing, Beijing Academy of Quantum Information Sciences, Beijing 100193, China}
\affiliation{Institute of Computing Technology, Chinese Academy of Sciences, Beijing 100190, China}
\affiliation{School of Physical Sciences, University of Chinese Academy of Sciences, Beijing 100190, China}
\author{Jintao Li}
\affiliation{Beijing Key Laboratory of Fault-Tolerant Quantum Computing, Beijing Academy of Quantum Information Sciences, Beijing 100193, China}
\author{Yipeng Zhang}
\affiliation{Beijing Key Laboratory of Fault-Tolerant Quantum Computing, Beijing Academy of Quantum Information Sciences, Beijing 100193, China}
\author{Bo Yang}
\affiliation{Beijing Key Laboratory of Fault-Tolerant Quantum Computing, Beijing Academy of Quantum Information Sciences, Beijing 100193, China}
\author{Xinkai Gao}
\affiliation{Longying Zhida (Beijing) Technology Co., Ltd., Beijing 100020, China}
\author{Xiaofeng Cao}
\affiliation{Longying Zhida (Beijing) Technology Co., Ltd., Beijing 100020, China}
\author{Kai Xu}
\affiliation{Beijing National Laboratory for Condensed Matter Physics, Institute of Physics, Chinese Academy of Sciences, Beijing 100190, China}
\affiliation{School of Physical Sciences, University of Chinese Academy of Sciences, Beijing 100190, China}
\affiliation{Beijing Key Laboratory of Fault-Tolerant Quantum Computing, Beijing Academy of Quantum Information Sciences, Beijing 100193, China}
\affiliation{Hefei National Laboratory, Hefei 230088, China}
\author{Pengpeng Hao}
\email{haopengpeng5709@sina.com}
\affiliation{Yale School of Management, Yale University, New Haven CT 06511, USA}
\author{Xuan Yang}
\email{yangxuan@lyzdfintech.com}
\affiliation{Longying Zhida (Beijing) Technology Co., Ltd., Beijing 100020, China}
\author{Heng Fan}
\email{hfan@iphy.ac.cn}
\affiliation{Beijing National Laboratory for Condensed Matter Physics, Institute of Physics, Chinese Academy of Sciences, Beijing 100190, China}
\affiliation{School of Physical Sciences, University of Chinese Academy of Sciences, Beijing 100190, China}
\affiliation{Beijing Key Laboratory of Fault-Tolerant Quantum Computing, Beijing Academy of Quantum Information Sciences, Beijing 100193, China}
\affiliation{Songshan Lake Materials Laboratory, Dongguan 523808, Guangdong, China}
\affiliation{Hefei National Laboratory, Hefei 230088, China}
\date{\today}
\begin{abstract}
Quantum Machine Learning (QML) offers a new paradigm for addressing complex financial problems intractable for classical methods. This work specifically tackles the challenge of few-shot credit risk assessment, a critical issue in inclusive finance where data scarcity and imbalance limit the effectiveness of conventional models. To address this, we design and implement a novel hybrid quantum-classical workflow. The methodology first employs an ensemble of classical machine learning models (Logistic Regression, Random Forest, XGBoost) for intelligent feature engineering and dimensionality reduction. Subsequently, a Quantum Neural Network (QNN), trained via the parameter-shift rule, serves as the core classifier. This framework was evaluated through numerical simulations and deployed on the Quafu Quantum Cloud Platform's ScQ-P21 superconducting processor. On a real-world credit dataset of 279 samples, our QNN achieved a robust average AUC of $0.852\pm 0.027$ in simulations and yielded an impressive AUC of $0.88$ in the hardware experiment. This performance surpasses a suite of classical benchmarks, with a particularly strong result on the recall metric. This study provides a pragmatic blueprint for applying quantum computing to data-constrained financial scenarios in the NISQ era and offers valuable empirical evidence supporting its potential in high-stakes applications like inclusive finance.
\end{abstract}
\maketitle
\end{CJK*}

\section{Introduction}

The field of quantum computing is undergoing remarkable progress, marked by an accelerated improvement in qubit counts and fidelities. The recent demonstration of small-scale quantum error correction~\cite{Acharya2024} suggests that the era of fault-tolerant quantum computing may be nearer than previously anticipated. This rapid advancement is prompting industries across various sectors to prepare for the quantum future, with the financial sector demonstrating particular interest.

This interest from the financial sector is twofold. First, a defensive motivation exists, driven by concerns that quantum computers could break the RSA encryption standard using Shor's algorithm~\cite{Shor1994}, thus spurring the search for post-quantum cryptography. Second, a proactive drive seeks to leverage the unique properties of quantum mechanics, such as superposition and entanglement, to solve computationally intractable problems in finance. Consequently, extensive research has been dedicated to applications in portfolio optimization, market prediction, asset pricing, and risk management~\cite{Gu2020, Nagel2021, Bagnara2022, Ta2018, Chen2021, Ban2018, Ma2021, Kumbure2022, Yuan2020, Rasekhschaffe2019}.

Broadly, these research efforts are advancing along two distinct paths. The first involves a long-term strategy focused on achieving significant speedups with future fault-tolerant quantum computers. For instance, Quantum Amplitude Estimation (QAE) promises a theoretical quadratic speedup over classical Monte Carlo methods, which is highly relevant for critical tasks like derivative pricing and Value at Risk (VaR/CVaR) calculations~\cite{Montanaro2015, Brassard2002, Woerner2019, Stamatopoulos2020}. Similarly, the HHL algorithm offers a potential exponential speedup for solving systems of linear equations, although its practical application faces stringent prerequisites~\cite{Harrow2009, Marfany2024}.

The second, more pragmatic path, focuses on developing heuristic solutions for current Noisy Intermediate-Scale Quantum (NISQ) devices. This category includes hybrid quantum-classical algorithms such as the Variational Quantum Eigensolver (VQE), the Quantum Approximate Optimization Algorithm (QAOA)~\cite{Farhi2014, Orus2019}, and Quantum Annealing (QA)~\cite{Venturelli2019}. Although widely applied to problems like portfolio optimization, the performance of these algorithms is severely constrained by hardware noise, limited circuit depth, and challenges in parameter optimization. Whether they can provide a tangible \textit{quantum advantage} over classical methods remains an open and actively debated question.

Beyond these established paths, quantum computing is poised to exert a transformative impact on machine learning (ML) within the financial sector. Quantum Machine Learning (QML) aims to harness quantum phenomena, such as superposition and entanglement, to accelerate classical ML algorithms or enhance their predictive power~\cite{Biamonte2017, Cerezo2022}. A variety of QML algorithms have been explored for financial applications.

Quantum Support Vector Machines (QSVMs) and their associated quantum kernel methods are among the most extensively studied QML techniques~\cite{Rebentrost2014, Herman2022}. These approaches employ a quantum feature map to project classical data into a high-dimensional Hilbert space. The underlying principle is that data points which are not linearly separable in their original space may become so in the quantum feature space, thereby simplifying complex classification tasks.

As the quantum analogues of classical neural networks, Quantum Neural Networks (QNNs) utilize Parameterized Quantum Circuits (PQCs) as their fundamental layers~\cite{Doosti2024}. The development of diverse QNN architectures including Quantum Convolutional Networks (QCNNs)~\cite{Cong2019}, Quantum Generative Adversarial Networks (QGANs)~\cite{Lloyd2018, Huang2021}, Quantum Circuit Born Machines (QCBMs)~\cite{Liu2018, Benedetti2019}, and Quantum Graph Neural Networks (QGNNs)~\cite{Verdon2019} underscores the potential for QML to process more complex and unstructured financial data.

Other notable QML algorithms include the quantum k-nearest neighbor, quantum k-means clustering, Quantum Principal Component Analysis (QPCA)~\cite{Lloyd2014}, and Quantum Reinforcement Learning (QRL)~\cite{Chen2020, Jerbi2021}. These methods are typically designed to accelerate specific computational bottlenecks within their classical counterparts, such as distance calculations or linear algebra subroutines.

The aforementioned QML algorithms hold significant promise for enhancing financial risk management. In credit scoring, for example, conventional models are often limited by their reliance on historical data and their difficulty in capturing complex, non-linear correlations. QML is well-positioned to enhance predictive accuracy by integrating a broader array of variables, including non-traditional data sources like social media activity and consumption patterns, and by uncovering intricate relationships that classical methods may overlook.

Similarly, in fraud detection, which is fundamentally an anomaly detection problem, QML offers a distinct advantage. The enhanced pattern-recognition capabilities of QML algorithms are particularly effective for analyzing vast streams of transaction data. This capability enables the real-time identification of subtle, suspicious patterns that might otherwise go unnoticed, thereby mitigating financial losses.

In this paper, we focus on the application of Quantum Neural Networks (QNNs) to few-shot learning for financial risk control. Within inclusive finance, the fundamental need for robust risk control persists, centered on constructing accurate models to assess customer risk effectively. However, traditional risk control methods face new challenges when applied to the large and diverse customer base of small and medium-sized enterprises (SMEs) and individual businesses. As customer groups are finely segmented, the sample size within each subgroup is often extremely small. This data scarcity is exacerbated by a severe lack of default instances ("bad samples"), creating a classic few-shot, imbalanced learning problem where conventional models struggle to achieve adequate generalization and precision.

While traditional modeling paradigms demand extensive data, QML offers a promising alternative for small-sample modeling in inclusive finance. The unique advantages of QML, such as its capacity for processing high-dimensional complex data and its potential for strong generalization, open new avenues for this domain. By leveraging quantum algorithms, financial institutions can potentially identify risks more accurately and optimize credit decisions, thereby expanding the coverage and depth of inclusive finance while ensuring fund security.

Recent theoretical work supports this direction. Caro et al.~\cite{Caro2022} demonstrated that QML models can exhibit strong generalization from limited training data, with a generalization error bound that scales favorably with the number of trainable parameters $T$ and the training dataset size $N$ as $O(\sqrt{T/N})$. Motivated by this finding, our work provides experimental validation of this principle in a practical financial context.

\section{Credit Risk Assessment}

The primary objective of a robust risk control model is the accurate identification and mitigation of potential credit risks. Traditionally, commercial banks have relied on rule-based models, which are typically constructed manually using historical data and domain-specific expertise. With the advent of artificial intelligence, these legacy systems are increasingly being supplemented or replaced by machine learning models that can learn patterns directly from data.

A critical prerequisite for both approaches, however, is the availability of extensive data to ensure strong generalization. This requirement becomes a significant bottleneck in scenarios involving fine-grained customer segmentation. In such niche segments, the available sample size is often limited, and instances of default ("bad samples") are exceedingly rare. These data-scarce and imbalanced environments severely undermine the ability of conventional models to achieve reliable generalization and high precision.

This context creates a pressing need for novel modeling techniques capable of strong generalization from limited data. The introduction of quantum algorithms, particularly from the field of Quantum Machine Learning (QML), is emerging as a promising solution to this challenge.

\subsection{Overview of the Scorecard Model}

The scorecard model is recognized as one of the most influential quantitative tools in the history of consumer and commercial credit assessment. It has long served as a cornerstone of risk management by converting complex customer data into a single, interpretable score. The primary objective of a scorecard is to quantitatively analyze customer attributes and behavioral data through a rigorous scientific modeling process and comprehensive model evaluation. This enables the effective assessment of credit risk or fraud risk and provides a solid foundation for informed credit decision-making.

Fundamentally, a scorecard is a binary classification model designed to estimate the probability of a customer defaulting, denoted as $P(Y=1|X)$, where $X$ represents the customer's feature such as age, income and occupation. For a certain type of customer, a corresponding model can be established by analyzing their historical data samples $D=\{(x_i,y_i) | x_i \in \mathbb{R}^n, y_i \in \{0,1\}\}$. Therefore, the quantity and quality of its samples directly affect the accuracy of the scorecard model.
These binary classifiers are broadly categorized into parametric models (e.g., logistic regression, support vector machines) and non-parametric models (e.g., tree-based models).

In practice, financial institutions commonly employ logistic regression for classification due to its interpretability. This method involves linear fitting historical data samples to estimate the probability of default. Suppose each customer group is characterized by a multivariate distribution of attributes $\textbf{X}_i$, which may include variables such as age, income, family size, credit history, and occupation. The corresponding decision boundary is as follows:
\begin{equation}
    \textbf{y} = \alpha + \beta_1 \textbf{X}_1 + \beta_2 \textbf{X}_2 + \dots + \beta_n \textbf{X}_n = \beta^{T}\textbf{X} + \alpha,
\end{equation}
when $y > 0$, it indicates that there is a high probability that the customer will default, otherwise, the probability is very low. Therefore, by introducing the Sigmoid function to represent this nonlinear relationship, the probability of default can be expressed as
\begin{equation}
    P(Y=1|X) = \frac{1}{1+e^{-\beta^{T} X - \alpha}}
\end{equation}

The objective is to determine an optimal set of model parameters $\alpha$ and $\beta$ by minimizing a loss function. A standard choice for binary classification tasks is the binary cross-entropy loss, which for a dataset of size $n$ is defined as:
\begin{equation}
	\begin{aligned}
		L(\alpha,\beta)=&\frac{1}{n} \sum_{i=1}^n\left[-y_i \log \left[P\left(y_i=1 \mid x_i, \alpha,\beta\right)\right]\right.
		\\
		& \left. -\left(1-y_i\right) \log \left[\left(1-P\left(y_i=1 \mid x_i, \alpha,\beta\right)\right)\right]\right] \\
		=&\frac{1}{n} \sum_{i=1}^n\left[-y_i(\beta^T x_i+\alpha)+\log(1+e^{\beta^T x_i + \alpha})\right],
	\end{aligned}
\end{equation}
where the parameters are typically optimized iteratively using an algorithm such as gradient descent. 

In contrast, non-parametric models like decision trees are constructed on a different principle. They build a tree-like topology by recursively partitioning the dataset. The splitting criterion at each node is determined by the feature that yields the highest Information Gain. Information Gain measures the reduction in uncertainty (or impurity), quantified by entropy, after a dataset is split on a particular attribute. It is calculated as:
\begin{equation}
	\mathrm{Gain}=E(S)-E_A(S),
\end{equation}
where $E(S)$ is the information entropy of the training set $S$. If $S$ contains $m$ classes with respective proportions $p_i(i=1,2,3,\dots,m)$, its entropy is defined as:
\begin{equation}
	E(S) = -\sum^m_{i=1}p_i \log(p_i).
\end{equation}

Here, $E_A(S)$ is the expected entropy after partitioning $S$ by attribute $A$. If attribute $A$ partitions the set $S$ into $k$ subsets $S_1, \dots, S_k$, the expected entropy is given by:
\begin{equation}
	E_A(S)=\sum^k_{i=1}\frac{|S_i|}{S} E(S_i).
\end{equation}
The attribute $A$ that maximizes the information gain is chosen as the branching node, as it provides the most information for classification.

\section{Quantum Neural Networks}

Quantum Neural Networks (QNNs) represent a central paradigm in Quantum Machine Learning (QML), harnessing fundamental quantum phenomena, namely superposition and entanglement, to perform advanced information processing.

A primary advantage of QNNs lies in their inherent quantum parallelism, which stems from the principle of superposition. A register of just $n$ qubits can represent a state in a $2^n$-dimensional Hilbert space, allowing for the simultaneous processing of an exponential number of classical states. This property offers significant potential for efficiently managing high-dimensional data. Furthermore, the phenomenon of entanglement allows QNNs to establish and leverage non-local correlations between qubits. This capability is crucial for representing complex data structures and functions that are intractable for classical models, with a prime example being the accurate modeling of intermolecular interactions in quantum chemistry.

Theoretical studies also suggest that for certain tasks (e.g., solving linear equations), QNNs may exhibit a lower model complexity than their classical counterparts, potentially offering advantages in navigating complex, non-convex optimization landscapes. The operational mechanism of modern QNNs often relies on variational quantum circuits. In this hybrid quantum-classical approach, a quantum circuit is equipped with trainable, parameterized gates whose parameters are optimized using classical algorithms, in a manner analogous to the training of classical neural networks. A typical QNN architecture is composed of the following key stages:

\subsection{Quantum Data Encoding Layer} 
Classical data needs to be transformed into quantum states through quantum encoding (e.g., basis encoding \cite{Farhi2018}, angle encoding, or amplitude encoding \cite{Zoufal2019} \cite{Wiebe2012} \cite{CarreraVazquez2021}). 
Quantum data embedding is the bridge between classical data and quantum computing, which is the process of converting classical data into quantum states that can be processed by quantum computers. This transformation is a crucial initial step in the vast majority of quantum algorithms, especially QML. The choice and efficiency of encoding methods have a significant impact on the performance and computational power of subsequent algorithms. By mapping data to a high-dimensional Hilbert space, combined with the ability to achieve nonlinear transformations through quantum gates, it provides the potential to capture complex and nonlinear relationships in data that may be intractable for classical models. This is very powerful for machine learning and one of the key motivations driving the development of quantum machine learning. There are currently several mainstream encoding methods:

Basis Encoding maps classical binary strings directly onto the computational ground state of quantum bit systems. Specifically, the classical bit $0$ is mapped to the quantum state $|0\rangle$, and the classical bit $1$ is mapped to the quantum state $|1\rangle$. The main limitation of ground state encoding is its high sensitivity to bit flip errors and its unsuitability for handling complex data structures compared to amplitude encoding or angle encoding. Its direct mapping relationship means that a bit flip error on a quantum bit will directly damage the corresponding classical bit information.

Amplitude Encoding is the process of encoding a classical $N-$dimensional data vector into the amplitudes of $n$ qubits. The generated quantum state is the superposition of the calculated ground state:
\begin{equation}
    |\psi_x\rangle = \sum_{i=1}x_i|i\rangle,
\end{equation}
where, $x_i$ is the $i$-th element of the classical vector, and $|i\rangle$ is the $i$-th computational basis . This method is highly efficient in the use of quantum bits, as it can encode $2n$ classical values with only $n$ quantum bits, fully utilizing the characteristic that quantum states can be in superposition states with complex amplitudes, providing exponential data compression, which is its main advantage. However, the circuit for preparing a quantum state with any specified amplitude can be very complex and deep. The implementation of amplitude encoding circuits is usually not trivial and requires arbitrary quantum state preparation technology. The depth of the encoded circuit is usually linearly related to the input size $N$, which may quickly become impractical on NISQ devices~\cite{Pagni2025}.

Angle encoding is the process of mapping the eigenvalues of classical data into angle parameters for quantum gate rotation operations. Usually, each classical feature $x_j$ controls the rotation angle of a rotation gate applied to the $j-$th qubit. It is common to directly set the angle of a single qubit rotation gate (such as $\mathrm{RX (x_j)}$ or $\mathrm{RY (x_j)}$) applied to the corresponding qubit. For example, if an RY gate is used, the state of the jth quantum bit becomes
\begin{equation}
    |\psi_i\rangle = \cos(x_i/2)|0\rangle +\sin(x_i/2)|1\rangle,
\end{equation}
In addition, there are other encoding methods such as Hamiltonian encoding~\cite{Mitra2003}, which are mainly used for quantum simulation problems. There is a clear trade-off between quantum bit efficiency and circuit complexity/depth in state preparation. Amplitude encoding has high quantum bit efficiency $O (\log N)$, but preparing arbitrary states with specific amplitudes may require deep and complex circuits. The quantum bit efficiency of ground state encoding and angle encoding is relatively low $O(N)$, but the encoding circuit of a single feature can be simpler/shallower, especially for simple angle encoding without entanglement. This trade-off is a crucial design choice, especially in the NISQ era where the number of qubits and circuit depth are limited.

\begin{figure*}[htbp]
    \centering 
    \includegraphics[width=0.9\textwidth]{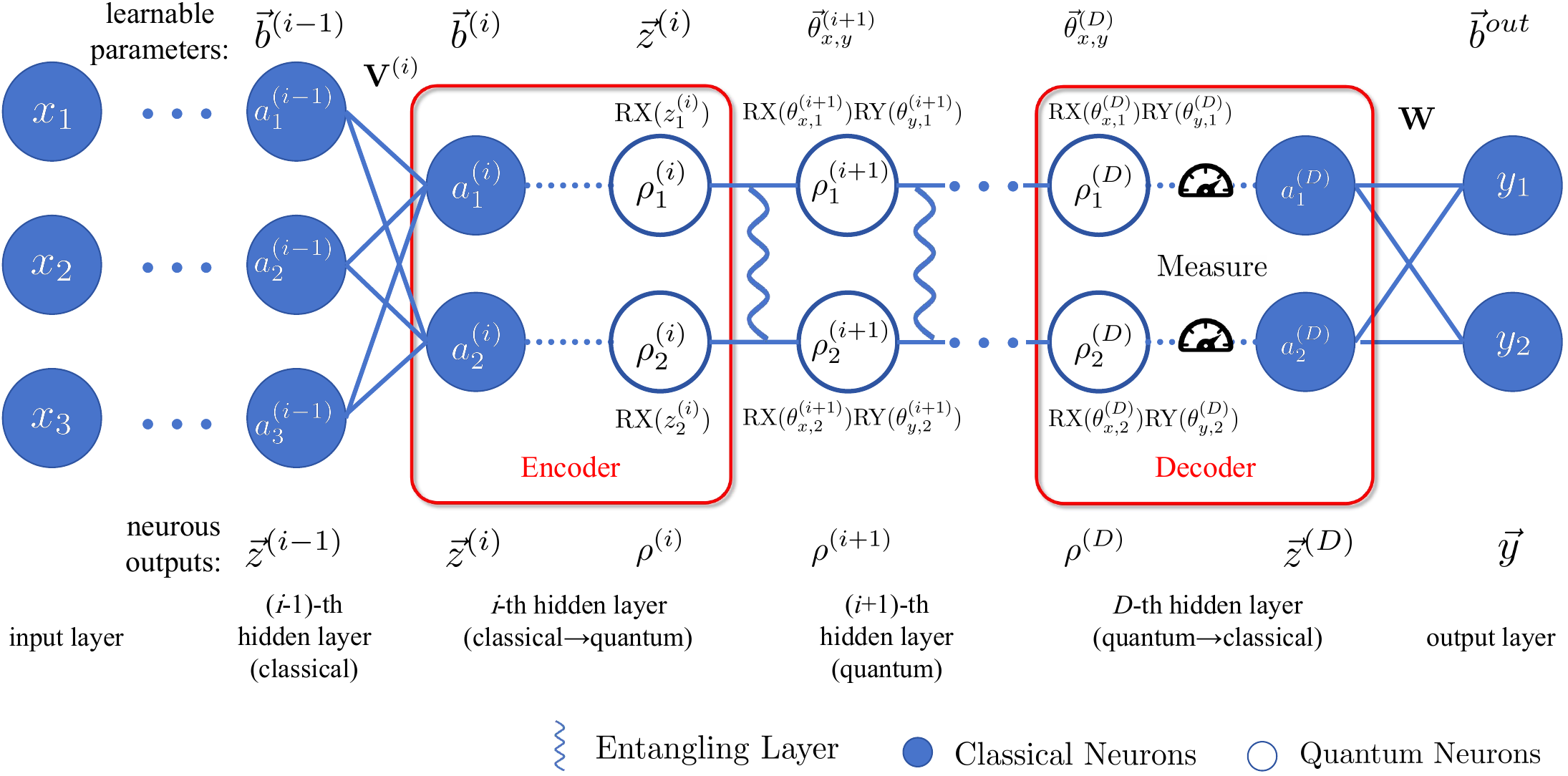}
    \caption{Quantum-classical hybrid neural networks and the schematic diagram of the hybrid backpropagation method. The blue circles are classical nodes, the white circles are quantum nodes, and the diagram also lists the inputs, outputs, and trainable parameters for each node.}
    \label{fig:hybridNN} 
\end{figure*}

\subsection{Parametrized Quantum Circuit Layer} 

In quantum neural networks, Parameterized Quantum Circuit (PQC), also known as Variational Quantum Circuit (VQC), is the core unit~\cite{Chen2025}. PQC is a special type of quantum circuit that contains a series of quantum gates, some of which are controlled by a set of adjustable, continuous classical parameters (such as the angle of a single qubit rotation gate). By adjusting these classical parameters, PQC can prepare various quantum states, effectively exploring a specific subspace in the vast Hilbert space. It is this parameterized characteristic that distinguishes PQC from standard quantum circuits with fixed gate operations. It transforms quantum circuits from a static algorithm into a trainable and optimizable model, which is conceptually very similar to classical machine learning models. This feature makes PQC the cornerstone of all variational quantum algorithms (VQA).
The calculation process of a PQC is to map a classical input data vector $\vec{x}$ and a parameter vector $\vec{\theta}$ to a classical output value~\cite{Tseng2025}. This output value is usually obtained by measuring the expected value of an observable $O$. The complete process is 
\begin{equation}
    f(\vec{x}, \vec{\theta})=\left\langle\psi_0\right| V^{\dagger}(\vec{x}) U^{\dagger}(\vec{\theta}) O U(\vec{\theta}) V(\vec{x})\left|\psi_0\right\rangle,
\end{equation}
where $|\psi_0\rangle$ is the initial state, $V(\vec{x})$ is the data encoding layer shown in the previous subsection, $O$ is the observable, and $U(\vec{\theta})$ is the PQC, which is the trainable engine in the model. Ansatz is a sequence composed of a series of quantum gates, namely $U(\vec{\theta})=U_L (\vec{\theta}_L) \dots U_1 (\vec{\theta}_1)$. These parameters are functionally similar to weights and biases in classical neural networks. The proposed structure, including the selected quantum gate types (such as single bit rotation gates and entanglement gates like CNOT) and their arrangement, collectively determine the computing power of PQC. How to design an efficient prototype is a core issue in current quantum computing research, which requires a trade-off between the model's expressive power and trainability.Expressive ability refers to the ability of PQC to generate diverse quantum states, that is, it can explore the range of Hilbert space. High expressiveness is ideal because it implies that the model has the potential to represent complex functions or accurately approximate the true solution of a problem.
The trainability refers to the difficulty level of optimizing PQC parameters.
There is a contradictory relationship between these two. Research has shown that proposals with strong expressive power, especially those with structures similar to random quantum circuits, are more prone to the phenomenon of "barren plateau", where gradients disappear exponentially with system size, resulting in the model being unable to train~\cite{Tangpanitanon2020}. How to find the best balance point in this dilemma is a core challenge in PQC design.

\subsection{Quantum Measurement} 

When measuring a quantum system in a superposition state, wave function collapse occurs. The measurement behavior will force the system to choose a definite state from its uncertain superposition state, and this process is instantaneous and irreversible.
More specifically, every observable physical quantity in quantum mechanics is represented by an Hermitian operator $A$. This operator has a set of eigenstates
$|\phi_i\rangle$ and its corresponding eigenvalue $a_i$ satisfy the eigenequation $A|\phi_i\rangle = a_i |\phi_i\rangle$. These eigenvalues represent all possible deterministic results obtained when measuring the physical quantity.
When a system is in any superposition state $|\psi\rangle$, it can be expanded into a linear combination of the eigenstates of operator $A$:
\begin{equation}
    |\psi\rangle =\sum_i c_i |\phi_i\rangle,
\end{equation}
where $c_i=\langle \phi_i|\psi \rangle$ is probability amplitude. According to the measurement postulate of quantum mechanics, when measuring the physical quantity A of the system, the measurement result must be one of the eigenvalues $a_i$ of operator $A$. The probability of obtaining the result $a_i$ is given by the Born rule, which means that the modulus squared of the corresponding probability amplitude $P(a_i)=|c_i|^2=|\langle\phi_i| \psi\rangle|^2$.
Once the measurement is completed and the result $a_i$ is obtained, the state vector of the system will immediately collapse from the original superposition state $|\psi\rangle$ to the corresponding eigenstate $|\phi_i\rangle$. This process is irreversibleis, it  can be represented as
\begin{equation}
    |\psi\rangle = \sum_j c_j |\phi_j\rangle \xrightarrow{\text{measured}\  a_i} |\psi'\rangle = |\phi_i\rangle.
\end{equation}

\subsection{Quantum Gradient and Optimizer} 
Parameter shift is an analytical method for accurately calculating the expected gradient of parameterized quantum circuits. Consider a quantum gate $U(\theta) =e^{-i\theta G/2}$ controlled by parameter $\theta$. To calculate the gradient, simply add and subtract a fixed displacement (such as $\pi/2$) from the original parameter values, run the quantum circuit twice to obtain two expected values, subtract them, and multiply them by a constant coefficient. See Appx.~\ref{appx:backpropagation} for details.
The standard binomial parameter shift is mainly applicable to single parameter quantum gates generated by Hermitian operators with two different eigenvalues, the most typical examples of which are Pauli rotation gates (such as $\text{RX}(\theta)$, $\text{RY}(\theta)$, $\text{RZ}(\theta)$) and phase gates $\text{P}(\theta)$.

However, actual quantum circuits may contain more complex gates, such as generators with two or more different eigenvalues or multi parameter gates. For these situations, the simple binomial parameter shift rule is no longer applicable, and generalized parameter shift or stochastic parameter shift need to be used~\cite{Wierichs2022}. Parameter shift typically exhibit better robustness in quantum systems with noise, using relatively large, fixed parameter displacements (e.g. $\pm\pi/2$ for Pauli rotation gates). This larger displacement usually results in larger changes in the expected value, making it less sensitive to small noise fluctuations.

\subsection{Quantum-classical hybrid neural networks}
Quantum-classical hybrid neural networks are also becoming popular. By embedding quantum neural networks in classical neural networks, classical neural network components and quantum computing units are synergistically combined. The core idea lies in making full use of the respective advantages of classical computing and quantum computing. Usually, a classical neural network is responsible for data preprocessing and/or post-processing of quantum results, while the quantum neural network layer exists as a special layer in the network, and its parameters are trained and optimized through the cyclic iteration of quantum gradients and classical optimizer. When necessary, both classical and quantum neural networks can participate in the entire backpropagation (BP), and quantum gradients can be connected with classical gradients as shown in Fig.~\ref{fig:hybridNN}. Firstly consider forward propagation (FP), which is the simpler case. It is known that in a classical neural network, the output of the first neuron in the $i$-th hidden layer is
\begin{equation}
    z^{(i)}_1 = \sigma(a^{(i)}_1) = \sigma(\sum_j v_{j1} z_j^{(i-1)}+b_1^{(i)}),
\end{equation}
which is equal to the dot product of the previous layer output $z^{(i-1)}_j$ and weight $v_{j1}$, with the bias $b^{(i)}_1$ and activation function $\sigma(\bullet)$. Assuming we introduce a quantum layer as the $(i+1)$-th layer, we need to encode the classical information $z^{(i)}$ output from the $i$-th layer into quantum information. A classical layer and a quantum layer appear simultaneously in the encoder (red box). However, they provide dual perspectives on the same set of parameters.
Then it propagates like a regular quantum neural network until it needs to be connected to the classical one again. At this point, it is necessary to decode quantum information into classical information, which is achieved through measurement. The decoder also includes a pair of mirrored neural layers. Finally, like classical neural network models, there is an output layer to calculate the loss function.
Backpropagation is slightly more complicated, which involves the transmission of quantum parameter and classical parameter gradients. For the quantum layer, we can still perform calculations through parameter shift, so the key to the entire propagation lies in the $i$-th layer where the Encoder is located. Suppose to calculate the gradient of a parameter $a^{(i-1)}_1$ which is in the $(i-1)$-th classical hidden layer, we have
\begin{equation}
    \label{aaa}
    \begin{aligned}
        &\frac{\partial{}}{\partial{a^{(i-1)}_1}}{Loss(\vec{y})} \\
        &= \sum_j \frac{\partial{Loss(\vec{y})}}{\partial{z_j^{(i)}}}
        \frac{\partial{z_j^{(i)}}}{\partial{a_j^{(i)}}}
        \frac{\partial{a_j^{(i)}}}{\partial{a_1^{(i-1)}}}\\
        &= \sum_j \frac{\partial{Loss(\vec{y})}}{\partial{z_j^{(i)}}}
        \frac{\partial{z_j^{(i)}}}{\partial{a_j^{(i)}}}
        \sum_k \frac{\partial({v_{k1} z_1^{(i-1)}+b_1^{(i)}})}{\partial{a_1^{(i-1)}}} \\
        &= \sum_j\sum_k \frac{\partial{Loss(\vec{y})}}{\partial{z_j^{(i)}}} \sigma^{\prime}(a_j^{(i)}) v_{k1} \sigma^{\prime}(a_1^{(i-1)}) \\
    \end{aligned}
\end{equation}

Simply obtaining the derivative of the parameters of the encoding operator in the encoder $\partial{Loss(\vec{y})}/\partial{z_j^{(i)}}$ can smoothly connect to the classical backpropagation chain rule, which can also be easily obtained through parameter shift. In this way, a complete forward and back propagation is completed.

The emergence of this hybrid approach provides a practical path for exploring potential quantum advantages on existing or recently achievable quantum devices, which may realize practical application value earlier than full quantum algorithms. Traditionally, quantum advantage typically refers to achieving exponential acceleration when solving a problem or solving problems that classical computers find difficult to handle (such as Shor's algorithm). 

However, quantum-classical hybrid means that the entire system is not purely quantum. Therefore, quantum advantage may not entail an end-to-end speedup, but rather manifest in other forms - such as improved model expressivity for certain data types, enhanced generalization on specific problem classes, or more efficient exploration of the quantum parameter space - even when overall runtime remains comparable.

\section{Methods and Results}

The dataset used in this study is an open-source, anonymized dataset simulating personal internet loans, originally comprising 10,000 samples across 14 industries. To ensure model specificity and account for sectoral variations in repayment ability, we focused our analysis on a sub-population of 279 borrowers from the retail and wholesale industries.

The final dataset contains eight explanatory variables: cumulative early repayment amount, loan interest rate, the lower and upper bounds of the borrower's credit score range, online loan level, length of employment, and two anonymized features. The response variable is a binary indicator of loan default ($Y=1$). Within this dataset, 41 samples are defaults, resulting in a class imbalance with a default rate of approximately $14.6\%$.

\begin{figure*}[htbp]
    \centering 
    \includegraphics[width=0.9\textwidth]{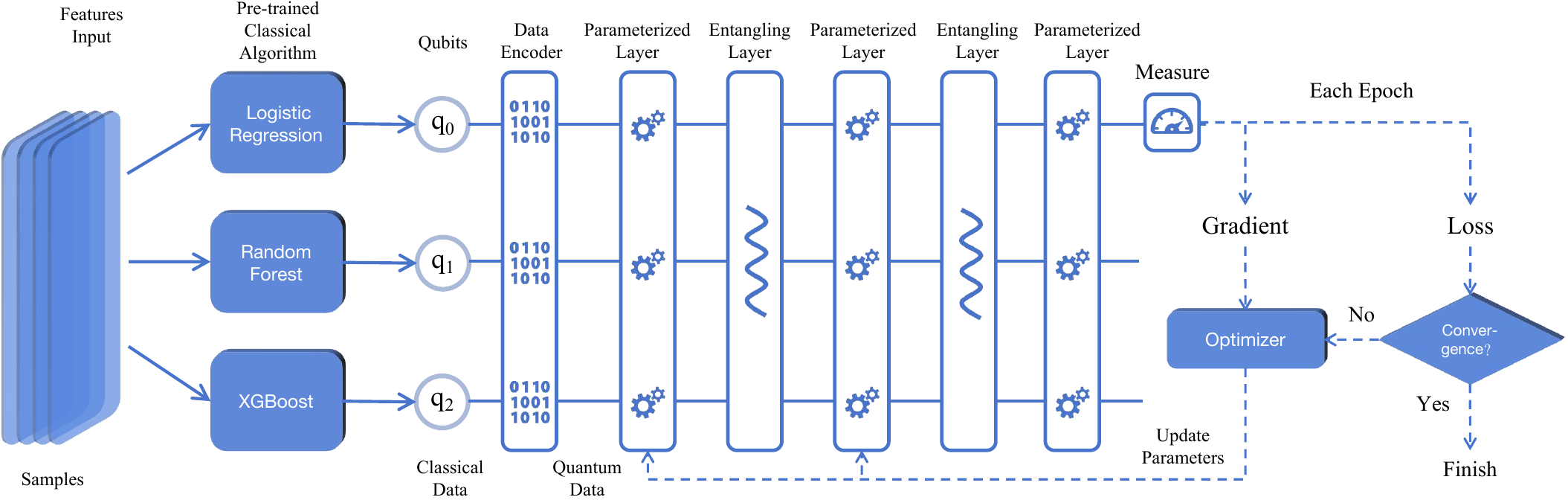}
    \caption{A demonstration of a hybrid neural network method with classic preprocessing. The blue circle represents the encoding of preprocessed classical feature data into quantum information. The quantum neural network consists of a sandwich structure consisting of parameter layers and entanglement layers, and updates parameters by solving quantum gradients and using classical optimizers}
    \label{fig:QNN2} 
\end{figure*}

To construct a compact yet informative feature set, we employed a model-based strategy for feature engineering and dimensionality reduction. This approach distills the original features into a dense representation that captures the most salient predictive information. Specifically, using a technique analogous to model stacking, we first trained three distinct models which are logistic regression, random forest, and XGBoost, on the original eight-dimensional data. The default probability predictions from these three models were then aggregated to form a new, three-dimensional feature vector for each sample. This compressed 3D representation served as the input for our quantum model. To make a fair comparison with the classical algorithm, the classical algorithm in the result (Tab.~\ref{tab:res-real}) also adopted the same preprocessing process.

Recognizing that the train-test split can significantly influence model performance, particularly with small and imbalanced datasets, we employed a repeated random sub-sampling cross-validation procedure. To ensure a robust evaluation, the entire dataset was partitioned 10 times. Each partition utilized a stratified split, maintaining a 7:3 ratio for both default and non-default samples across the training and testing sets. The detailed results of this procedure are presented in Tab.~\ref{tab:my_label}.


\begin{table*}
    \centering
    \begin{tabular}{c c c c c c c c c c c}
         \hline 
         Partition No. & 1 & 2 & 3 & 4 & 5 & 6 & 7 & 8 & 9 & 10\\
         \hline 
         Training Samples & 195 & 195 & 195 & 195 & 195 & 195 & 195 & 195 & 195 & 195 \\
         Default in Train & 29 & 29 & 29 & 29 & 29 & 29 & 29 & 29 & 29 & 29 \\
         Testing Samples & 84 & 84 & 84 & 84 & 84 & 84 & 84 & 84 & 84 & 84 \\
         Default in Test & 12 & 12 & 12 & 12 & 12 & 12 & 12 & 12 & 12 & 12 \\
         Batch Size & 64 & 64 & 32 & 32 & 32 & 64 & 128 & 128 & 32 & 32 \\
         \hline 
    \end{tabular}
    \caption{Detailed information of the cross-validation partitions}
    \label{tab:my_label}
\end{table*}

The model training was conducted using the \texttt{PyQuafu} software toolkit, which facilitated both simulations and experiments on physical hardware. As a core component of the Quafu quantum computing cloud platform, \texttt{PyQuafu} provides unified access to various backends, including local simulators, cloud-based simulators, and real quantum processors.

A key feature for this work is its seamless integration with the \texttt{PyTorch} framework, where a quantum circuit can be directly implemented as a differentiable neural network module (\texttt{torch.nn.Module()}). This enables automatic differentiation and backpropagation for variational quantum parameters (see App.~\ref{appx:backpropagation}) and greatly simplifies the construction and training of hybrid quantum-classical models. Within this framework, we employed a binary cross-entropy loss function (\texttt{torch.nn.BCELoss()}) and the AdamW optimizer (\texttt{torch.optim.AdamW()}) for model training.

\subsection{Numerical Result}

The architecture of our QNN model, depicted in Fig.~\ref{fig:QNN2}, is composed of three primary stages. First, the input features are encoded into the quantum state using a layer of single-qubit RX gates. The subsequent Parameterized Quantum Circuit (PQC) consists of an alternating sequence of three parameterized layers and two entangling layers. The entangling layers form a CNOT chain with periodic boundary conditions, while each parameterized layer contains trainable $\text{RX}(\vec{\theta}_{l,i,x})$ and $\text{RY}(\vec{\theta}_{l,i,y})$ gates. In the final parameterized layer, a modification was made such that only the gate parameters applied to the first qubit were optimized. Finally, only the first qubit is measured, which is a deliberate design choice intended to mitigate the barren plateau problem and enhance trainability.

\begin{figure}[htbp]
    \centering 
    \includegraphics[width=0.5\textwidth]{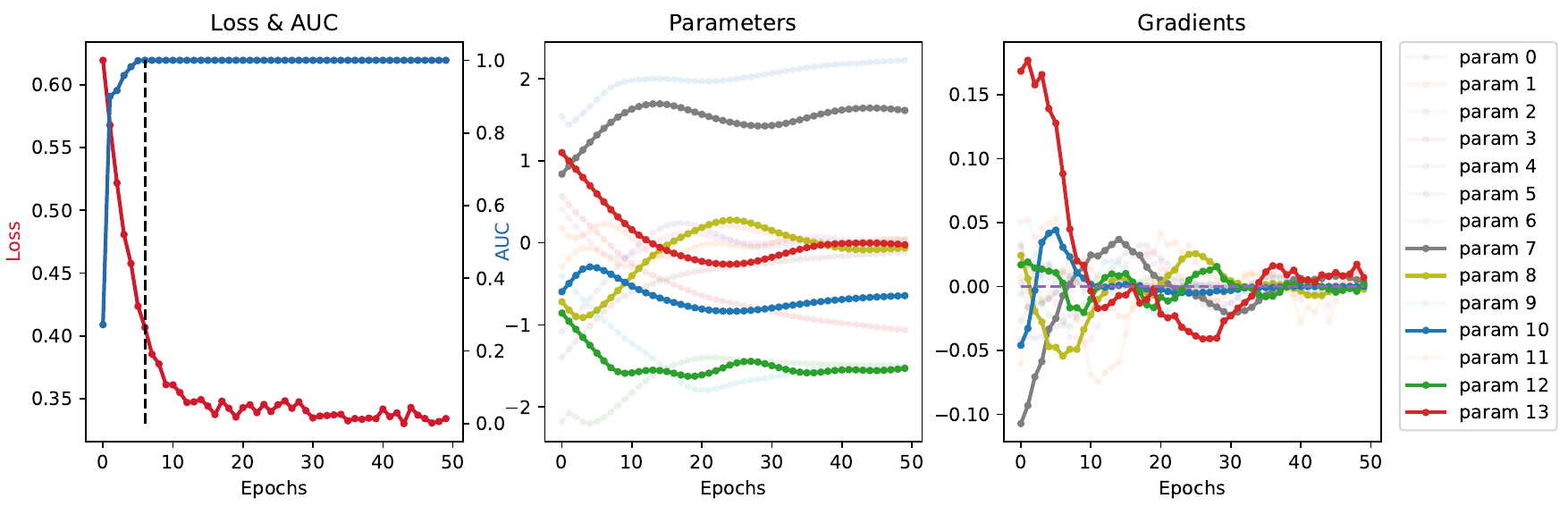}
    \caption{Training dynamics for Partition $8$. The left panel shows the evolution of the loss function (red) and AUC (blue). The middle and right panels track the trajectories of the $14$ QNN model parameters and their corresponding gradients respectively, demonstrating model convergence. For visual clarity, the trajectories of five representative parameters are highlighted.}
    \label{fig:loss}
\end{figure}

To illustrate the training dynamics, we present the results for Partition $8$ as a representative case. The model was trained for $50$ epochs on the simulator, with key metrics summarized in Fig.~\ref{fig:loss}. Effective training was confirmed by the convergence of the loss function, model parameters, and their gradients. As shown, the loss steadily decreased and stabilized around the $20$th epoch, with the parameters exhibiting a similar convergence trend. Concurrently, the parameter gradients diminished towards zero with oscillations.

\begin{figure*}[htbp]
    \centering 
    \includegraphics[width=1.0\textwidth]{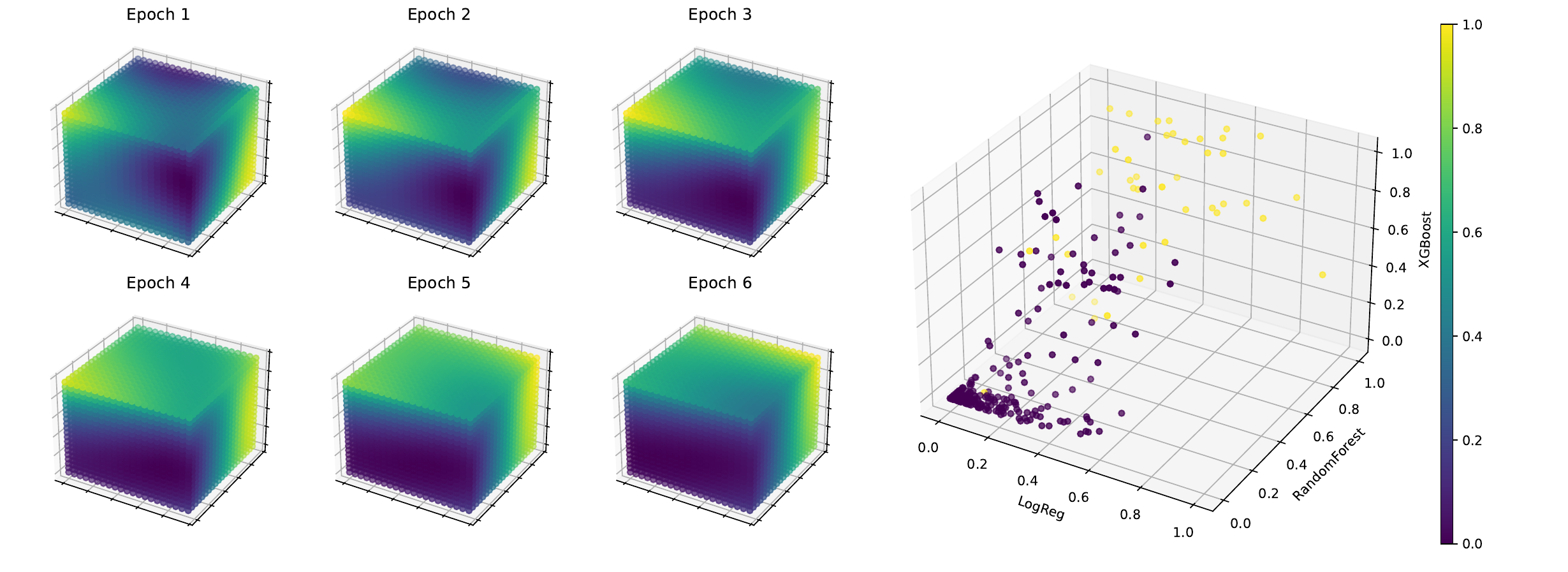}
    \caption{The results of numerical simulation using a quantum computing simulator. The left figure shows the evolution of the model's predicted distribution values with increasing epochs, while the right figure shows the overall distribution of the real small sample data. It can be clearly seen that as the training progresses, the distribution of the model's predicted values gradually approaches the true values.}
    \label{fig:sim-res-scatter}
\end{figure*}

Notably, the Area Under the Curve (AUC) on the training set increased rapidly, reaching its maximum value by the $6$th epoch. This indicates that the model had achieved perfect separation of the training data, and any further training posed a significant risk of overfitting. Consequently, to ensure generalization, we selected the model parameters from the 6th epoch for the final model. Fig.~\ref{fig:sim-res-scatter} provides a visual assessment of this process. It illustrates how the distribution of the model's predicted outputs evolves over the first six epochs. Starting from a random distribution due to untrained parameters, the predictions are shown to gradually align with the true sample distribution (depicted in the rightmost panel).


\begin{figure}[htbp]
    \centering 
    \includegraphics[width=0.5\textwidth]{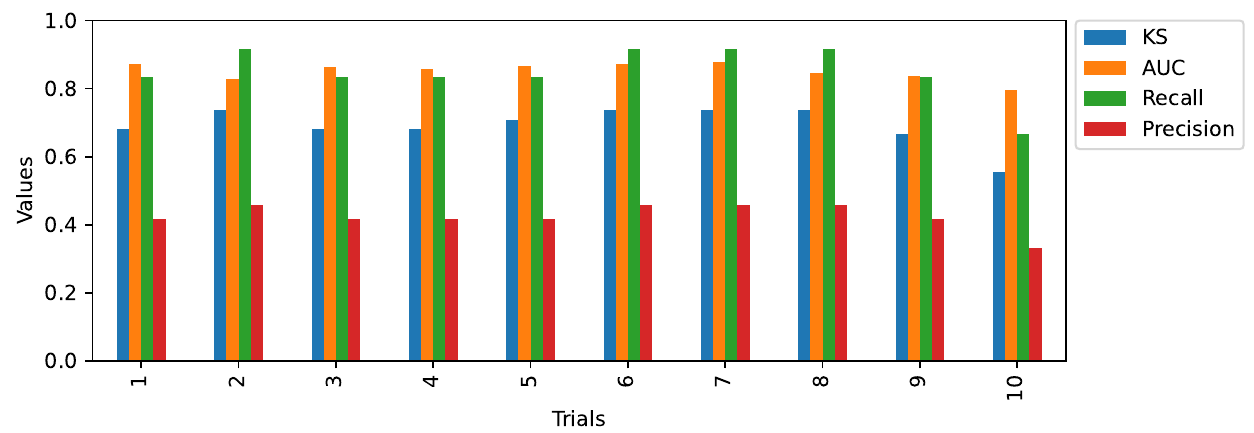}\textbf{}
    \caption{Robustness of the model in numerical simulations, confirmed via 10-run cross-validation. This bar chart displays the model's final performance metric (AUC, KS, recall and precision) for 10 trials, each utilizing a distinct, stratified random split of the dataset. The consistent height of the bars demonstrates the model's stability against variations in data partitioning.}
    \label{fig:sim-res-bar}
\end{figure}

The trained model was evaluated on the test set, where it demonstrated robust and competitive performance. Specifically, our QNN model achieved an average AUC of $0.853 \pm 0.025$ across the test sets in simulation. This level of performance shows a distinct improvement over a suite of classical benchmarks, including both traditional and machine learning algorithms. Tab.~\ref{tab:res-real} summarizes the full suite of average performance metrics across the $10$ experimental 
trials, while Fig.~\ref{fig:sim-res-bar} provides a detailed breakdown for each individual trial. Furthermore, the model's high stability across these different trial indicates its strong generalization capability.


\begin{table}
    \centering
    \begin{tabular}{c c c c c}
        \hline
        Algorithm & AUC & KS & Recall & Precision \\ [0.5ex]
        \hline
        Logistic Regression & 0.844 & 0.686 & 0.850 & 0.408 \\
        Decision Tree & 0.706 & 0.447 & 0.567 & 0.377 \\
        Random Forest & 0.817 & 0.640 & 0.750 & 0.360 \\
        SVM & 0.702 & 0.501 & 0.658 & 0.316 \\
        XGBoost & 0.806 & 0.603 & 0.717 & 0.349 \\
        Neural Networks & 0.82 & 0.62 & 0.74 & 0.37 \\
        QNN(Simulator) & 0.853 & 0.692 & 0.850 & 0.425 \\
        QNN(ScQ-P21) & 0.88 & 0.73 & 0.88 & 0.44\\ [0.5ex]
        \hline
    \end{tabular}
    \caption{Comparison of results between quantum neural networks run on quantum computing cloud platforms and classical methods}
    \label{tab:res-real}
\end{table}

\subsection{Experimental Result}


To bridge the gap between theoretical performance and real-world applicability, we transitioned from simulation to validate our findings on physical quantum hardware. This crucial step is essential for assessing the practical viability of quantum algorithms in the current NISQ era. The experiments were executed on the ScQ-P21, a superconducting quantum chip accessible via the Quafu Quantum Computing Cloud Platform. Full device specifications and calibration data, which informed our architectural decisions, are provided in Appx.~\ref{appx:devic}.


\begin{figure}[htbp]
	\subfigure[Experimental QNN circuit architecture.]
	{
		\includegraphics[scale=0.2]{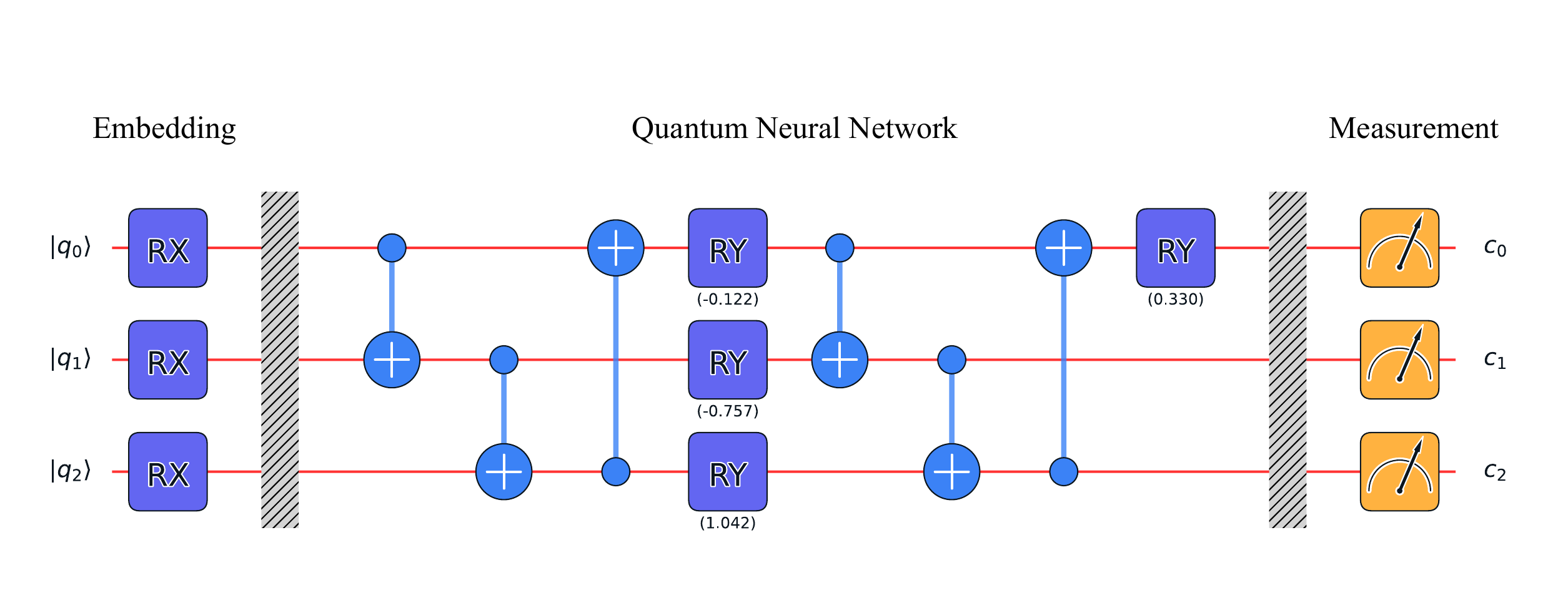}
	}
	\subfigure[Model training on the ScQ-P21 quantum processor.] 
	{
		\includegraphics[scale=0.6]{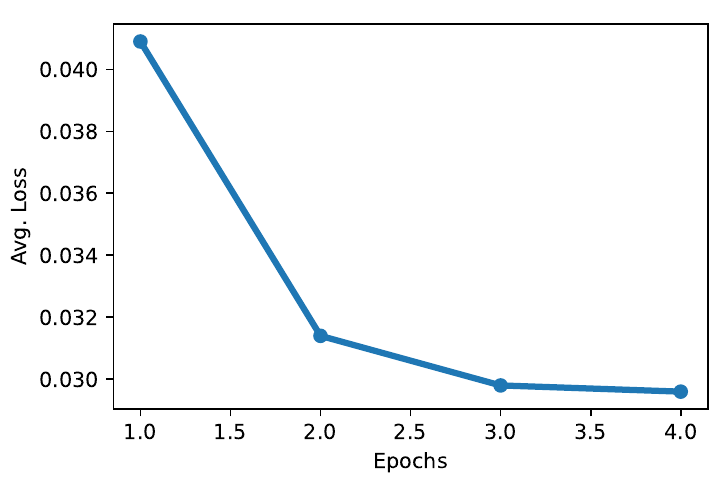}
	}
	\caption{Experimental validation of the QNN model on the Quafu quantum cloud platform. (a) The simplified quantum circuit architecture used for the hardware experiment. The circuit depth and number of parameters were reduced compared to the simulation to accommodate hardware noise. (b) The corresponding training process on the physical hardware. The plot tracks the loss function across training epochs, showing a clear convergence trend that indicates successful model optimization.}
\end{figure}

Executing complex quantum circuits on NISQ devices presents significant challenges, including limited gate fidelities, short qubit coherence times, and restricted connectivity. To accommodate these hardware constraints and mitigate the impact of noise, the QNN architecture used in simulations was strategically simplified. Specifically, the parameterized layers were constructed using only RY rotation gates, which are often among the native gates with lower error rates on superconducting hardware. Furthermore, the circuit depth was substantially reduced to two parameterized and two entangling layers, as minimizing the number of gate operations is a primary strategy for enhancing performance under noisy conditions. The exact hardware-adapted circuit is depicted in Fig.~\ref{fig:fulu1_chip}.

Remarkably, despite this simplified architecture and the inherent hardware noise, the experiment yielded an impressive AUC of 0.88. This result not only surpasses the average performance from our noiseless simulations (though it falls within one standard deviation) but also provides valuable insights into QML performance on real devices. We hypothesize two potential factors contributing to this strong outcome. First, the model may have benefited from a particularly favorable random parameter initialization. Second, it is plausible that the inherent noise of the quantum device acted as a form of regularization, preventing the model from overfitting to the small training set and thus improving its generalization on the test set which is a phenomenon sometimes observed in machine learning.

This finding underscores a key principle of NISQ-era QML: simpler, hardware-aware models can be highly expressive and may even outperform more complex theoretical designs. It suggests not only potential parameter redundancy in our simulated model but also a promising direction for future research in hardware-software co-design for financial applications.



\section{Conclusion}



In this study, we successfully demonstrated the practical application of a hybrid quantum-classical neural network for few-shot credit risk assessment, a critical challenge in data-scarce financial environments. Our approach combines a classical machine learning ensemble for intelligent feature engineering with a quantum neural network classifier. The efficacy of this framework was validated through both simulation and physical hardware experiments, achieving a robust average AUC of $0.853 \pm 0.025$ in simulation and an impressive AUC of $0.88$ on a superconducting quantum processor. This performance not only surpasses a suite of classical benchmarks but also achieves a high recall rate, which is paramount for effective risk management.

Beyond the performance metrics, our work offers crucial insights for the NISQ era. Notably, our simplified, hardware-adapted QNN outperformed its more complex simulated counterpart, suggesting that inherent device noise may act as a beneficial regularizer and that simpler, hardware-aware models can be exceptionally powerful. This finding provides a pragmatic blueprint for near-term quantum applications and underscores the importance of hardware-software co-design. While our results are promising, future work should aim to validate this approach across more diverse datasets to establish broader generalizability. Continued progress will also depend on advancements in quantum hardware to mitigate noise and on further theoretical investigation into the generalization properties of QML models, a discussion to which our empirical results contribute.

\addcontentsline{toc}{chapter}{Acknowledgment}
\section*{Acknowledgments}
This work was supported by Beijing Municipal Science and Technology Commission, Administrative Commission of Zhongguancun Science Park (Grant No. Z231100001323001), National Natural Science Foundation of China (Grants Nos. 92265207, 
12247168 
), China Postdoctoral Science Foundation (Certificate Number: 2022TQ0036).
Quafu Quantum Computing Cloud Platform (https://quafu.baqis.ac.cn) is supported by Beijing Academy of Quantum Information Sciences (BAQIS), the Institute of Physics (IOP) of the Chinese Academy of Sciences (CAS), and the Synergetic Extreme Condition User Facility (SECUF) in Huairou District, Beijing.

\bibliography{quantum_finance}

\appendix
\section{Evaluation indicators for binary risk control models} 
In binary classification model evaluation, the Confusion Matrix is used to summarize the accuracy of the classification model's prediction results. The Confusion Matrix divides the prediction results into four quadrants, corresponding to True Positive (TP), False Positive (FP), True Negative (TN), and False Negative (FN), respectively

The four indicators in the table are explained as follows:
True example (TP): The number of samples that are actually positive classes and predicted by the model to be positive classes.
False positive examples (FP): The number of samples that are actually negative but predicted by the model as positive, also known as false positives.
True Counterexample (TN): The number of samples that are actually negative classes and predicted by the model to be negative classes.
False Counterexample (FN): The number of samples that are actually positive but predicted by the model as negative, also known as false negatives.
Based on the confusion matrix, multiple evaluation metrics can be calculated, such as AUC, KS, Recall, and Precision. The following is an introduction to these metrics.

\begin{table}[htbp]
    \begin{tabular}{c c c}
        \hline
         & Predicted Positive & Predicted Negative \\ [0.5ex]
        \hline
        Actually Positive & TP & FN \\
        Actually Negative & FP & TN \\ [0.5ex]
        \hline
    \end{tabular}
    \caption{title}
\end{table}

The area under the ROC curve (AUC) represents the probability that the model, if given a randomly chosen positive and negative example, will rank the positive higher than the negative. By setting different thresholds, the corresponding true case rate (TPR) and false positive case rate (FPR) are calculated, and the corresponding points are connected to form an ROC curve. AUC, which is the area below the ROC curve, is used to measure the overall performance of the binary classification model. The calculation formula is:
\begin{equation}
    \mathrm{AUC}=\frac{1}{P \times N}\sum(p_i,n_i)_{p_i > n_i}
\end{equation}

Where, $P$ and $N$ represent the number of positive and negative samples respectively, $p_i$ and $n_i$ represent the positive and negative sample prediction score respectively. The larger the AUC value, the better the performance of the model.

The Kolmogorov-Smirnov (KS) value is the maximum difference between the true rate (TPR) and the false positive rate (FPR) as a function of the threshold. That is
\begin{equation}
    \mathrm{KS} = \max(\mathrm{TPR}-\mathrm{FPR})
\end{equation}

Where, $\mathrm{TPR}=\mathrm{TP}/(\mathrm{TP}+\mathrm{FN})$ and $\mathrm{FPR}=\mathrm{FP}/(\mathrm{TN}+\mathrm{FN})$. The KS value can measure the model's ability to distinguish between positive and negative samples. The larger the KS value, the stronger the model's ability to distinguish between positive and negative samples.

Recall is used to measure the classifier's ability to identify positive samples.
\begin{equation}
    \mathrm{Recall}=\mathrm{TP}/(\mathrm{TP}+\mathrm{FN})
\end{equation}

The higher the recall rate, the more positive samples the classifier finds, but it may be accompanied by more negative samples being misjudged as positive samples.

Precision is used to measure the proportion of instances predicted by a classifier as positive samples that are truly positive.
\begin{equation}
    \mathrm{Precision}=\mathrm{TP}/(\mathrm{TP}+\mathrm{FP})
\end{equation}

The higher the precision, the higher the proportion of instances predicted as positive by the classifier that are truly positive, but some positive samples may be missed.

\section{Backpropagation Method for Quantum Neural Network\label{appx:backpropagation}}
In this section, we elaborate the implementation of quantum neural network (QNN) module in \texttt{PyQuafu} based on \texttt{PyTorch}~\cite{ansel2024}.
Unlike classical machine learning, where gradient calculations are performed on classical hardware (such as CPUs, GPUs, or ASICs) using backpropagation supported by automatic differenciation~\cite{baydin2017},
QNNs leverage the \emph{parameter-shift rule}~\cite{Wierichs2022} to derive gradients.
We illustrate the principle of parameter-shift as follows.
Consider a quantum circuit is parameterized by $n$ parameters $\theta = [\theta_1,\theta_2,\dots,\theta_n]$, the expectation value of measurements of certain observables can be expressed as a function,

\begin{equation}
    f\left( \theta \right)  = \langle \psi_i | U_i^\dagger\left( \theta_i \right)U_{i+1:n}^\dagger \hat{O} U_{n:i+1}U_i\left( \theta_i \right) |\psi_i \rangle,
\end{equation}

\noindent
where $\theta_i$ is the scalar parameter whose gradient is to be calculated, and $U_i\left( \theta_i \right) $ is the gate where $\theta_i$ lies in.
For notation simplicity, we have already absorbed the unitaries before $U_i\left( \theta_i \right)$ into $|\psi_i\rangle$, thus $|\psi_i\rangle = U_{i-1:1} |0\rangle = \prod_{i-1}^1 U_k |0\rangle$.
Assume $U_i(\theta_i) = e^{-\frac{i}{2}\theta_i P_j}$, where $P_j$ is the Pauli matrix.
We have,
\begin{equation}
    \label{eq:ps_rule}
	\frac{\partial f\left( \theta_i \right) }{\partial \theta_i} = \frac{1}{2} \left( f\left( \theta; \theta_i + \frac{\pi}{2}  \right) - f\left( \theta; \theta_i - \frac{\pi}{2} \right) \right).
\end{equation}

By inserting $\pm\pi/2$ rotation gates and measuring the expectation values respectively, we can calculate the exact gradient.

\begin{algorithm}[htbp]
    \SetAlgoLined
    \SetKwFunction{len}{len}
    \SetKwFunction{identity}{identity}
    \SetKwFunction{zeros}{zeros}
    \SetKwFunction{tolist}{tolist}
    \SetKwFunction{concat}{concat}
    \SetKwFunction{GenerateParameterShiftValues}{generate\_parameter\_shift\_values}
    \SetKwFunction{CalcGrad}{calculate\_gradients}
    \SetKwFunction{EstExp}{estimate\_expection}
    \SetKwInOut{KwIn}{Input}
    \SetKwInOut{KwOut}{Output}
    \SetKwProg{Fn}{Function}{}{end}

    \KwIn{ $\theta$ \tcp{Vector of parameters in the quantum circuit}}
    \KwOut{ $L_{\theta}$ \tcp{List of shifted parameter vectors}}
    \Fn{\GenerateParameterShiftValues{$\theta$}}{
        $n$ = \len{ $\theta$ } \tcp{Get the number of parameters}
        $\mathbf{I}$ = \identity($n$) \tcp{Create an identity matrix of size $n$}

        $L_{\theta_+}$ = $\theta$ + $\mathbf{I} \times \frac{\pi}{2}$ \tcp{Calculate positive parameter shifts}
        $L_{\theta_-}$ = $\theta$ - $\mathbf{I} \times \frac{\pi}{2}$ \tcp{Calculate negative parameter shifts}

    $L_{\theta}$ = \concat{$L_{\theta_+},L_{\theta_-}$} \tcp{Combine both lists of shifted parameters}

        \Return{ $L_{\theta}$ }
    }

    \KwIn{ $\mathcal{O}$ \tcp{The observable to be measured}}
    \KwIn{ $\theta$ \tcp{Vector of parameters in the quantum circuit}}
    \KwOut{ $\nabla_{\theta}$ \tcp{Gradients of parameters}}
    \Fn{\CalcGrad{$\mathcal{O}$,$\theta$}}{
        $L_{\theta}$ = \GenerateParameterShiftValues{$\theta$} \tcp{Generate shifted parameter vectors}
        $k$ = \len{$L_{\theta}$}
        $r$ = \zeros{$k$} \tcp{Initialize gradients as a vector with $k$ elements}
        \For{$i \gets 1$ \KwTo $k$}{
            $\theta \gets L_{\theta}[i]$ \tcp{Obtain the $i$th shifted parameter vector}
            $r[i] \gets$ \EstExp{$\mathcal{O},\theta$} \tcp{Estimate expectation value of measuring observable $\mathcal{O}$ using QPUs}
        }
        $\nabla_{\theta} \gets \frac{r[:\frac{k}{2}] - r[\frac{k}{2}:]}{2} $ \tcp{Calculate gradients based on the parameter-shift rule}
        \Return{$\nabla_{\theta} $}
    }

    \caption{Quantum gradient calculation based on the parameter-shift rule.}
    \label{alg:param_shift}
\end{algorithm}

\begin{table*}[htbp]
\begin{tabular}{c c c c c c c c c c}
\hline
Qubit & $Q_2$ & $Q_3$ & $Q_4$ & $Q_5$ & $Q_6$ & $Q_7$ & $Q_8$ & $Q_9$ & $Q_{11}$ \\ [0.5ex]
\hline
$T_1(us)$ & 32.6 & 25.0 & 26.2 & 33.7 & 22.4 & 21.3 & 33.5 & 31.6 & 32.7 \\
$T_2(us)$ & 12.0 & 4.0 & 8.1 & 19.4 & 11.7 & 15.7 & 10.1 & 16.1 & 8.9\\
$\eta / 2\pi$ (GHz) & 0.198 & 0.201 & 0.200 & 0.200 & 0.198 & 0.200 & 0.205 & 0.198 & 0.200 \\
$\omega^i/ 2\pi$ (GHz) & 5.130 & 4.889 & 5.052 & 5.024 & 5.059 & 4.894 & 5.167 & 5.145 & 4.886\\
$\omega^r/ 2\pi$ (GHz) & 6.771 & 6.84565 & 6.8101 & 6.9212 & 6.7871 & 6.8625 & 6.8297 & 6.9446 & 6.9193 \\
$F_0$ & 0.962 & 0.939 & 0.979 & 0.987 & 0.951 & 0.963 & 0.986 & 0.993 & 0.942 \\
$F_1$ & 0.932 & 0.820 & 0.891 & 0.930 & 0.909 & 0.873 & 0.932 & 0.929 & 0.883\\ 
Connection & $Q_2$-$Q_3$ & $Q_3$-$Q_4$ & $Q_3$-$Q_5$ & $Q_5$-$Q_7$ & $Q_6$-$Q_7$ & $Q_7$-$Q_8$ & $Q_7$-$Q_9$ & $Q_9$-$Q_{11}$ & -\\
$T_{CZ}$ (ns) & 47.8 & 41.5 & 39.6 & 39.5 & 40.0 & 40.9 & 39.1 & 40.5 & -\\
$F_{CZ}$ (ns) & 0.996 & 0.9947 & 0.9943 & 0.9954 & 0.9959 & 0.9953 & 0.9929 & 0.9951 & -\\[0.5ex]
\hline
\end{tabular}
\caption{Device information}
\label{tab:device}
\end{table*}

Based on the above principle, we build a \texttt{PyQuafu} module to interface with \texttt{PyTorch} by inheriting \texttt{autograd.Function} and \texttt{nn.Module}.
Specifically, we override the \texttt{backward} API by defining a subroutine for calculating the vector-jacobian product based on the parameter-shift rule.
The implementation of gradient calculation procedure is described by the pseudo-code in Alg.~\ref{alg:param_shift}.
Firstly, we generate the shifted parameter vectors $L_{\theta}$ by the function \texttt{generate\_parameter\_shift\_values}, in which $L_{\theta_+}$ and $L_{\theta_-}$ are given by,
\begin{equation}
    \label{eq:impl_theta_plus_and_minus}
    \begin{aligned}
        L_{\theta_+} &= \begin{bmatrix}
        \theta_1+\frac{\pi}{2} & \theta_1 & \cdots & \theta_1 \\
        \theta_2 & \theta_2 +  \frac{\pi}{2} & \cdots & \theta_2 \\
        \vdots & \vdots & \ddots & \vdots \\
        \theta_n & \theta_n & \cdots & \theta_n + \frac{\pi}{2}
    \end{bmatrix} \\
        L_{\theta_+} &= \begin{bmatrix}
        \theta_1-\frac{\pi}{2} & \theta_1 & \cdots & \theta_1 \\
        \theta_2 & \theta_2 -  \frac{\pi}{2} & \cdots & \theta_2 \\
        \vdots & \vdots & \ddots & \vdots \\
        \theta_n & \theta_n & \cdots & \theta_n - \frac{\pi}{2}
    \end{bmatrix}
    \end{aligned}
\end{equation}

Then we utilize the QPUs of the Quafu cloud to estimate the expectation value for each vector of shifted parameters and store the results in $r$ (line 11$\sim$ 13 in Alg.~\ref{alg:param_shift}).
In line with our illustrations about parameter-shift theory, we assume $\theta$ to be an $n$-dimensional vector, thus the generated $L_{\theta}$ encompasses $2n$ vectors with the first $n$ being $L_{\theta_+}$ and the second $n$ being $L_{\theta_-}$.
According to Eq.~\ref{eq:impl_theta_plus_and_minus} and Eq.~\ref{eq:ps_rule}, we see that $\nabla_{\theta_i} = (r[i] - r[n+i])/2$, thus we can calculate all the gradients as shown in line 14.

\begin{figure}[htbp]
\centering 
\includegraphics[width=0.45\textwidth]{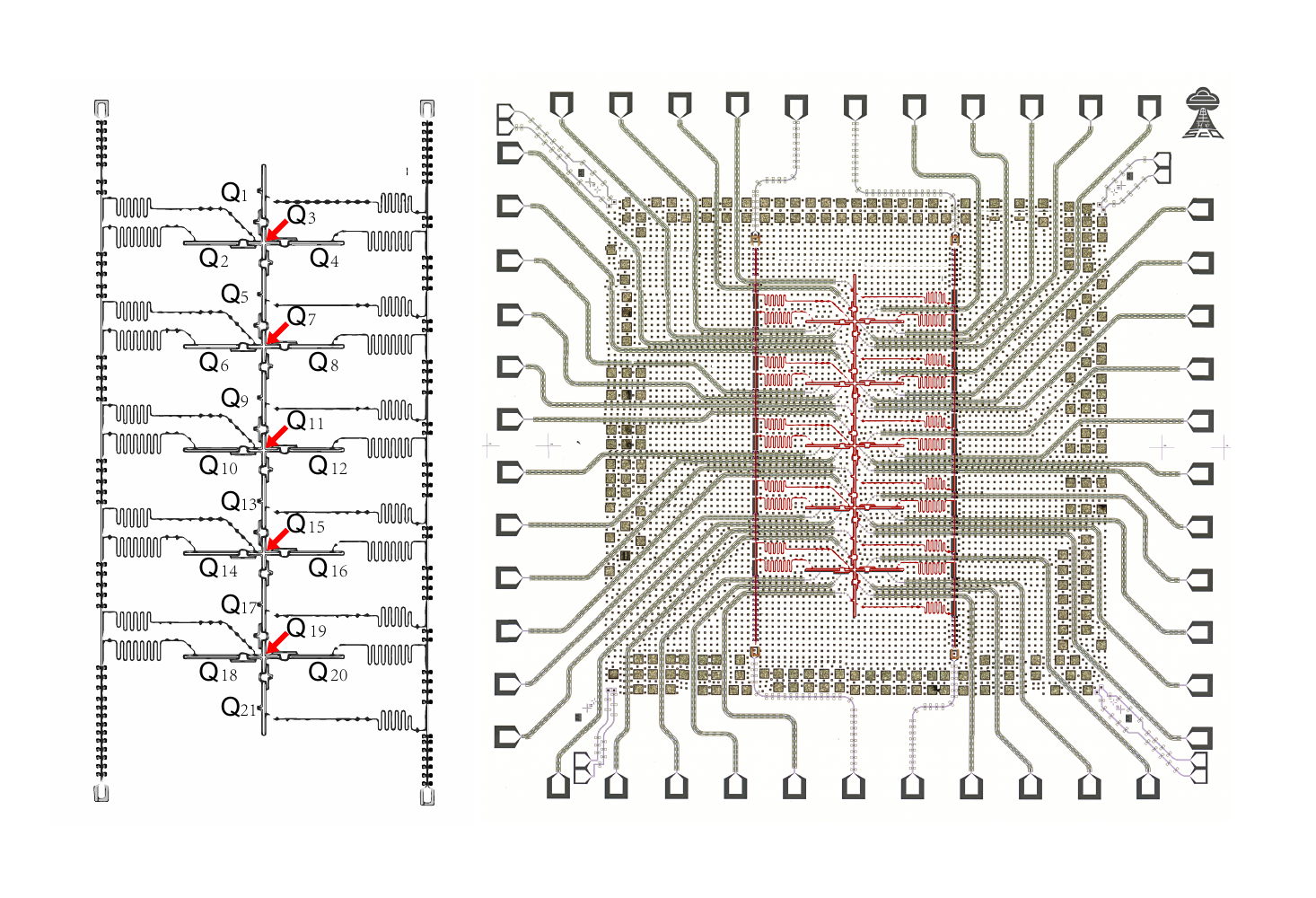}
\caption{Topological structure of the  21-qubits processor. The red layer represents the upper chip layer. Qubits ($Q_1$ to $Q_{21}$) are arrayed as shown in the left sub-figure.} 
\label{fig:fulu1_chip} 
\end{figure}

Finally, the gradients of parameters with respect to the loss $\mathcal{L}$ of the complete model are calculated by chain-rule,
i.e., $\nabla_{\theta} \mathcal{L}\left( \theta  \right) = \frac{\partial \mathcal{L}}{\partial f} \cdot \frac{\partial f}{\partial \theta}$,
where $\frac{\partial \mathcal{L}}{\partial f}$ is automatically calculated and provided by the \texttt{PyTorch} engine as it involves only classical tensor operations.
We encapsulate all these above steps in the customized \texttt{backward} API of the Quafu QNN module, which facilitates the training and testing of user-defined QNN models on the Quafu cloud quantum computers.

\section{Device Information and Quafu Quantum Cloud Platform\label{appx:devic}}

Quafu is an open cloud platform for quantum computation[Beijing Academy of Quantum Information Sciences, Institute of Physics of the Chinese Academy of Sciences, Tsinghua University, Quafu Quantum Cloud Computing Cluster, https://quafu.baqis.ac.cn/ (2024).], which provides several specifications of superconducting quantum processors. In our work, we use a 21-qubits processor of Quafu. This processor contains 20 tunable qubits and 19 tunable couplers integrated with flip-chip technology. Each qubit is tunable and dispersively couples with a resonator for single-shot readout. The qubits are coupled with a coupler between each of the two nearestneighbor qubit pairs. The topological structure of the qubits is shown in Fig.~\ref{fig:fulu1_chip}. 

In our work, we use a sub gate set which includes qubits from $Q_2$ to $Q_{11}$ and the couplers between them. The qubits parameters , coherence performance and CZ-gate fidelity can be found in the Table~\ref{tab:device}.

\end{document}